\documentclass{article} 
\usepackage{nips13submit_e,times}
\usepackage{url}

\usepackage{epsfig}
\usepackage{graphicx}
\usepackage{amsmath}
\usepackage{amssymb}
\usepackage{array}
\usepackage{algorithm,algorithmicx,algpseudocode,url,verbatim}

\usepackage[pagebackref=true,breaklinks=true,colorlinks,bookmarks=false,urlcolor=blue]{hyperref}

\title{Understanding Deep Architectures using a \\ Recursive Convolutional Network}

\author{
David Eigen \quad
Jason Rolfe \quad
Rob Fergus \quad
Yann LeCun
\\
Dept. of Computer Science, Courant Institute, New York University
\\
{\tt \{deigen,rolfe,fergus,yann\}@cs.nyu.edu}
}

\newcommand{\fig}[1]{Fig.~\ref{fig:#1}}
\newcommand{\tab}[1]{Table~\ref{tab:#1}}
\newcommand{\secc}[1]{Section~\ref{sec:#1}}

\def\etal{{\textit{et~al.~}}}

\newlength{\plotwidth}

\nipsfinalcopy 

\begin{document}

\maketitle

\begin{abstract}

A key challenge in designing convolutional network models is sizing them
appropriately.  Many factors are involved in these decisions, including number
of layers, feature maps, kernel sizes, etc.  Complicating this further
is the fact that each of these influence not only the numbers and dimensions of
the activation units, but also the total number of parameters.  In this paper
we focus on assessing the independent contributions of three of these linked
variables:  The numbers of layers, feature maps, and parameters.  To accomplish
this, we employ a recursive convolutional network whose weights are tied
between layers; this allows us to vary each of the three factors in a
controlled setting.  We find that while increasing the numbers of layers and
parameters each have clear benefit, the number of feature maps (and hence
dimensionality of the representation) appears ancillary, and finds most
of its benefit through the introduction of more weights.  Our results ({\it i})
empirically confirm the notion that adding layers alone increases computational
power, within the context of convolutional layers, and ({\it ii}) suggest that
precise sizing of convolutional feature map dimensions is itself of little
concern; more attention should be paid to the number of parameters in these
layers instead.

\end{abstract}

\section{Introduction}
\label{sec:intro}
\vspace{-2mm}

Convolutional networks have recently made significant progress in a variety of
image classification and detection tasks \cite{Kriz12,Ciresan11,Sermanet14},
with further gains and applications continuing to be realized.  At the same
time, performance of these models is determined by many interrelated
architectural factors; among these are the number of layers, feature map dimension,
spatial kernel extents, number of parameters, and pooling sizes and placement.
Notably, multiple layers of unpooled
convolution \cite{Kriz12,ZeilerArx13} have been utilized lately with considerable success.  These architectures must be carefully
designed and sized using good intuition along with extensive trial-and-error
experiments on a validation set.  But are there any characteristics to
convolutional layers' performance that might be used to help
inform such choices?  In this paper, we focus on disentangling and assessing
the independent effects of three important variables: the numbers of layers,
feature maps per layer, and parameters.

We accomplish this via a series of three experiments using
a novel type of recursive network model.  This model has a
convolutional architecture and is equivalent to a deep convolutional
network where all layers have the same number of feature maps and the
filters (weights) are tied across layers.
By aligning the architecture of this model to existing
convolutional approaches, we are able to tease apart these three factors that
determine performance.  For example, adding another layer increases the
number of parameters, but it also puts an additional non-linearity into the
system.  But would the extra parameters be better used expanding the size
of the existing layers? To provide a general answer to this type of
issue is difficult since multiple factors are conflated: the
capacity of the model (and of each layer) and its degree of non-linearity. However, we
can design a recursive model to have the same number of layers and
parameters as the standard convolutional model, and thereby see if the number
of feature maps (which differs) is important or not. Or we can match
the number of feature maps and parameters to see if the number of
layers (and number of non-linearities) matters. 

We consider convolutional models exclusively in this paper for several reasons.
First, these models are widely used for image recognition tasks.  Second, they
are often large, making architecture search tedious.  Third, and most
significantly, recent gains have been found by using stacks of multiple
unpooled convolution layers.  For example, the convolutional model proposed by
Krizhevsky \etal \cite{Kriz12} for ImageNet classification has five
convolutional layers which turn out to be key to its performance. In
\cite{ZeilerArx13}, Zeiler and Fergus reimplemented this model
and adjusted different parts in turn. One of the largests
effects came from changing two convolutional layers in the middle of the model:
removing them resulted in a 4.9\% drop in performance, while expanding them
improved performance by 3.0\%.  By comparison, removing the top two densely
connected layers yielded a 4.3\% drop, and expanding them a 1.7\% gain, even
though they have far more parameters. Hence the use of multiple convolutional
layers is vital and the development of superior models relies on understanding
their properties.  Our experiments have particular bearing
in characterizing these layers.

\subsection{Related Work}
\vspace{-1mm}
\label{sec:related}

The model we employ has relations to recurrent neural networks.
These are are well-studied models \cite{Hochreiter97,Schmid07,Sutskever10},
naturally suited to temporal and sequential data. For example, they have
recently been shown to deliver excellent performance for phoneme recognition
\cite{Graves13} and cursive handwriting recognition \cite{Graves09}.  However,
they have seen limited use on image data.  Socher \etal \cite{Socher11} showed
how image segments could be recursively merged to perform scene parsing. More
recently \cite{Socher12}, they used a convolutional network in a separate
stage to first learn features on RGB-Depth data, prior to hierarchical merging.
In these models the input dimension is twice that of the output. This
contrasts with our model which has the same input and output dimension.

Our network also has links to several auto-encoder models.  Sparse
coding \cite{Olshausen1997} uses iterative algorithms, such as ISTA
\cite{Beck09}, to perform inference. Rozell \etal \cite{Rozell08}
showed how the ISTA scheme can be unwrapped into a repeated series of
network layers, which can be viewed as a recursive net. Gregor \&
LeCun \cite{Gregor10} showed how to backpropagate through such a
network to give fast approximations to sparse coding known as LISTA. Rolfe \& LeCun
\cite{Rolfe13} then showed in their DrSAE model how a discriminative
term can be added.  Our network can be considered a purely
discriminative, convolutional version of LISTA or DrSAE.

There also are interesting relationships with convolutional Deep Belief
Networks \cite{Lee09}, as well as Multi-Prediction Deep Boltzmann Machines
\cite{Goodfellow13}.  As pointed out by \cite{Goodfellow13}, mean field
inference in such models can be unrolled and viewed as a type of recurrent
network.  In contrast to the model we use, however, \cite{Lee09} trains
unsupervised using contrastive divergence, while \cite{Goodfellow13} is
nonconvolutional and focuses on conditioning on random combinations of inputs
and targets.

\section{Approach}
\vspace{-1mm}

Our investigation is based on a multilayer Convolutional Network
\cite{MNIST}, for which all layers beyond the first have the same size and
connection topology.  All layers use rectified linear units (ReLU)
\cite{Coates2011, Glorot2011, Nair2010}.
We perform max-pooling with
non-overlapping windows after the first layer convolutions and rectification;
however, layers after the first use
no explicit pooling.  We refer to the number of feature maps per layer as $M$,
and the number of layers after the first as $L$.  To emphasize the difference between the
pooled first layer and the unpooled higher layers, we denote the first
convolution kernel by $\mathbf{V}$ and the kernels of the higher layers by
$\mathbf{W}^l$.   A per-map bias $\mathbf{b}^l$ is applied in conjunction with
the convolutions.  A final classification matrix $\mathbf{C}$ maps the last
hidden layer to softmax inputs.

\begin{figure}[tbh]
  \begin{center}
    \includegraphics[width=5.5in, keepaspectratio=true]{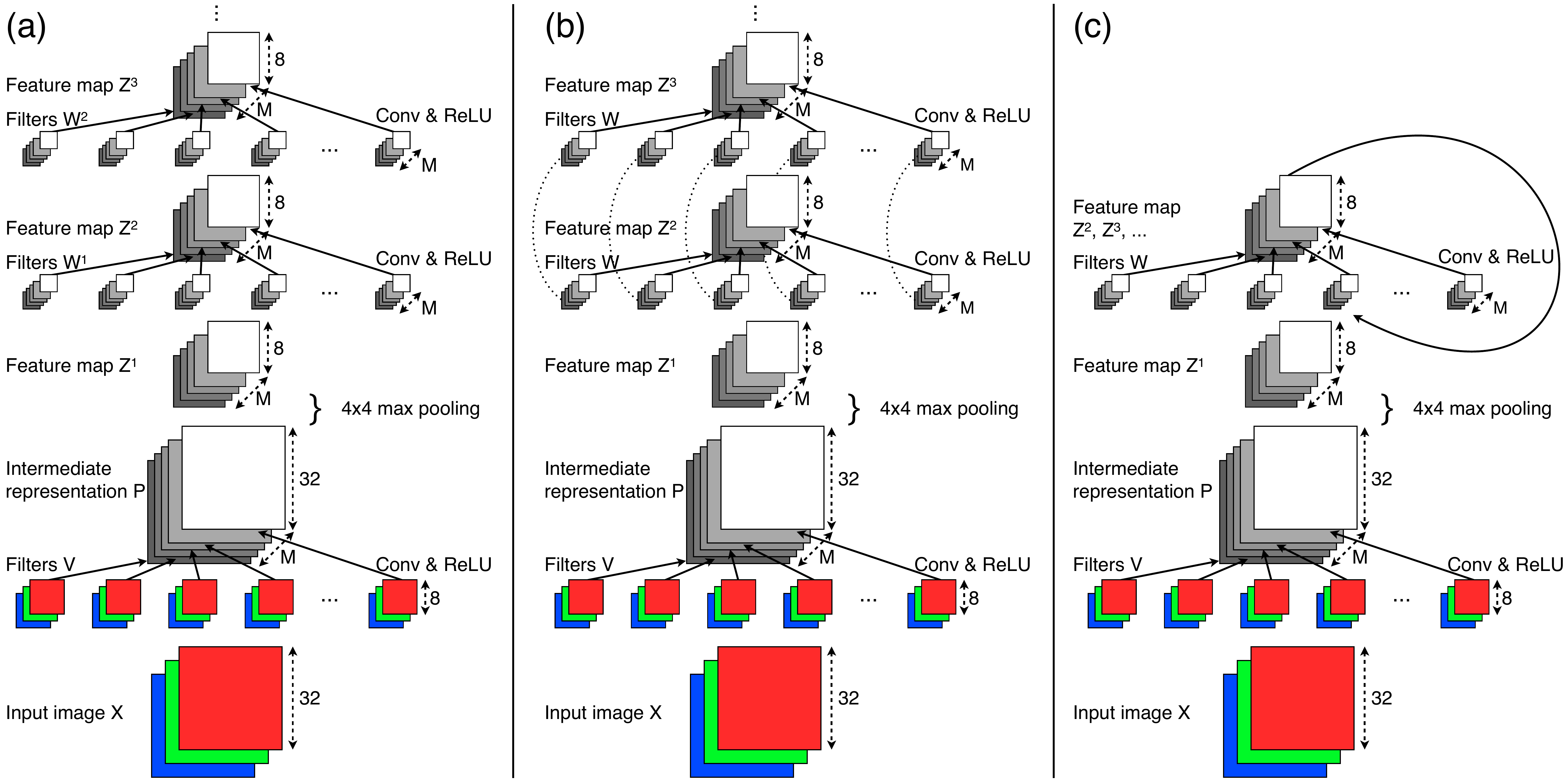}
  \end{center}
  \caption{Our model architecture prior to the classification layer, as applied to CIFAR and SVHN
    datasets. (a): Version with un-tied weights in the upper
    layers. (b): Version with tied weights. Kernels connected by dotted lines are constrained to be identical. (c): The network with tied
    weights from (b) can be represented as a recursive network. \label{fig:architecture_figure}} 
\end{figure}

Since all hidden layers have the same size, the transformations at all layers beyond
the first have the same number of parameters (and the same connection
topology).  In addition to the case where all layers are independently
parameterized, we consider networks for which the parameters of the higher
layers are tied between layers, so that $\mathbf{W}^i =
\mathbf{W^j}$ and $\mathbf{b}^i = \mathbf{b}^j$ for all $i,j$.  As shown in
\fig{architecture_figure}, tying the parameters across layers
renders the deep network dynamics equivalent to recurrence:  rather than
projecting through a stack of distinct layers, the hidden representation is
repeatedly processed by a consistent nonlinear transformation.  
The convolutional nature of the transformation performed at each layer implies
another set of ties, enforcing translation-invariance among the parameters.
This novel recursive, convolutional architecture
is reminiscent of LISTA \cite{Gregor10}, but without a direct projection from
the input to each hidden layer.

\subsection{Instantiation on CIFAR-10 and SVHN}

We describe our models for the CIFAR-10 \cite{Kriz09} and SVHN
\cite{SVHN} datasets used in our experiments. In both cases, each
image $\mathbf{X}^n$ is of size $32 \times 32 \times 3$.  In the
equations below, we drop the superscript $n$ indicating the index in
the dataset for notational simplicity.  The first layer applies a set
of $M$ kernels $\mathbf{V}_m$ of size $8 \times 8 \times 3$ via
spatial convolution with stride one (denoted as $*$), and per-map bias
$\mathbf{b}_m^0$, followed by the element-wise rectification
nonlinearity.  We use a ``same'' convolution (i.e. zero-padding
the edges), yielding a same-size representation $\mathbf{P}$ of $32 \times 32
\times M$.
This representation is then max-pooled within each feature map with
non-overlapping $4 \times 4$ windows, producing a hidden layer $\mathbf{Z^1}$ of size $8
\times 8 \times M$. 
\begin{equation*}
\mathbf{P}_{m} = \max\left(0, \mathbf{b}_{m}^0 + \mathbf{V}_m * \mathbf{X} \right)
~~ , \quad
\mathbf{Z}_{i,j,m}^1 = \max_{i',j' \in \left\{0, \cdots, 3 \right\}} \left(\mathbf{P}_{4\cdot i + i', 4 \cdot j + j', m} \right) 
\vspace{-1mm}
\end{equation*}
All $L$ succeeding hidden layers maintain this size, applying a set of $M$
kernels $\mathbf{W}_m^l$ of size $3 \times 3 \times M$, also via ``same'' spatial
convolution with stride one, and per-map bias $\mathbf{b}_m^l$, followed by the
rectification nonlinearity:  
\begin{equation*}
\mathbf{Z}_{m}^l = \max\left(0, \mathbf{b}_{m}^{l-1} + \mathbf{W}_m^{l-1} * \mathbf{Z}^{l-1} \right)
\end{equation*}
In the case of the tied model (see \fig{architecture_figure}(b)), the
kernels $W^l$ (and biases $b^l$) are constrained to be the same.  The
final hidden layer is subject to pixel-wise L2 normalization and
passed into a logistic classifier to produce a prediction $\mathbf{Y}$:
\begin{equation*}
\mathbf{Y}_k = \frac{\exp(Y'_k)}{\sum_k \exp(Y'_k)} \qquad \text{where} \qquad \mathbf{Y}'_k = \sum_{i,j,m} \mathbf{C}_{i,j,m}^k \cdot
{\mathbf{Z}_{i,j,m}^{L+1}} / ||\mathbf{Z}_{i,j}^{L+1}||
\vspace{-2mm}
\end{equation*}

\label{sec:training_details}
The first-layer kernels $\mathbf{V}_m$ are initialized from a zero-mean
Gaussian distribution with standard deviation $0.1$ for CIFAR-10 and $0.001$ for
SVHN.  The kernels of the higher layers $\mathbf{W}_m^l$ are initialized to the identity transformation $\mathbf{W}_{i',j',m',m} = \delta_{i',0} \cdot \delta_{j',0} \cdot \delta_{m',m}$, where $\delta$ is the Kronecker delta.  The network is trained to minimize the logistic loss function $\mathcal{L} = \sum_n 
\log(\mathbf{Y}_{k(n)}^n)$ and $k(n)$ is the true class of the
$n$th element of the dataset.  The parameters are not subject to explicit regularization.  Training is performed via stochastic gradient descent with
minibatches of size $128$, learning rate $10^{-3}$, and momentum~$0.9$:
$$
\mathbf{g} = 0.9 \cdot \mathbf{g} + \sum_{n \in \text{minibatch}} \frac{\partial \mathcal{L}^n}{\partial \left\{ \mathbf{V}, \mathbf{W}, \mathbf{b} \right\} } 
\quad ; \quad
\left\{ \mathbf{V}, \mathbf{W}, \mathbf{b} \right\} = \left\{ \mathbf{V}, \mathbf{W}, \mathbf{b} \right\} - 10^{-3} \cdot \mathbf{g}
$$

\section{Experiments}

\vspace{-2mm}
\subsection{Performance Evaluation}
\vspace{-1mm}

\setlength{\plotwidth}{2.8in}

We first provide an overview of the model's performance at different sizes, with both untied and tied weights, in order to examine basic
trends and compare with other current systems.  For \mbox{CIFAR-10}, we tested
the models using $M = 32, 64, 128, \text{or } 256$ feature maps per layer, and
$L = 1, 2, 4, 8, \text{or } 16$ layers beyond the first.  For SVHN, we used $M
= 32, 64, 128, \text{or } 256$ feature maps and $L = 1, 2, 4, \text{or } 8$
layers beyond the first.  That we were able to train networks at these large
depths is due to the initialization of all $W^l_m$ to the identity: this
initially copies activations at the first layer up to the last layer, and
gradients from the last layer to the first.  Both untied and tied models had
trouble learning with zero-centered Gaussian initializations at
some of the larger depths.

\begin{figure}[h]
\centering
\hspace{-1.5in}
\includegraphics[width=\plotwidth]{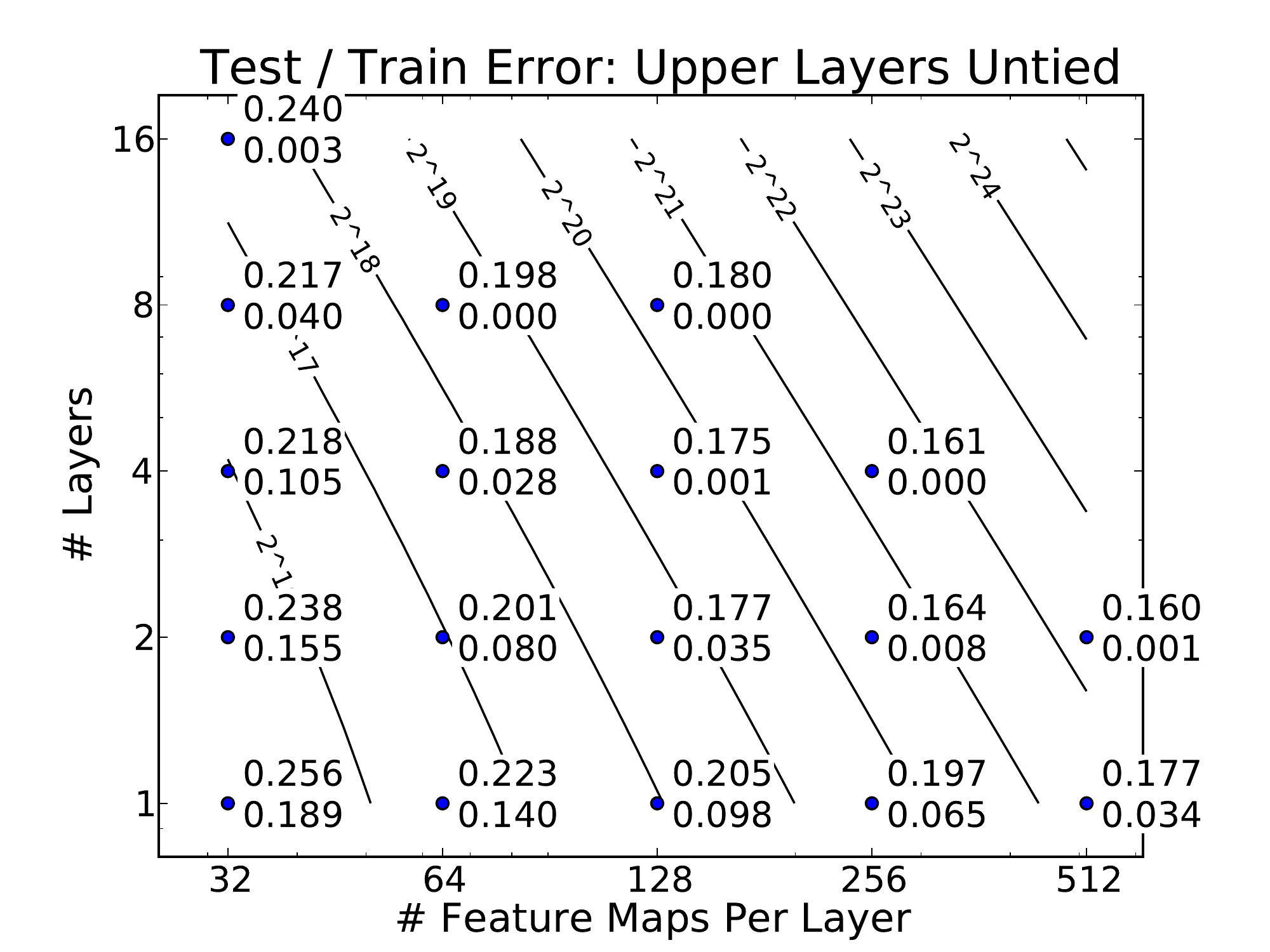}
\hspace{-0.1 in}
\includegraphics[width=\plotwidth]{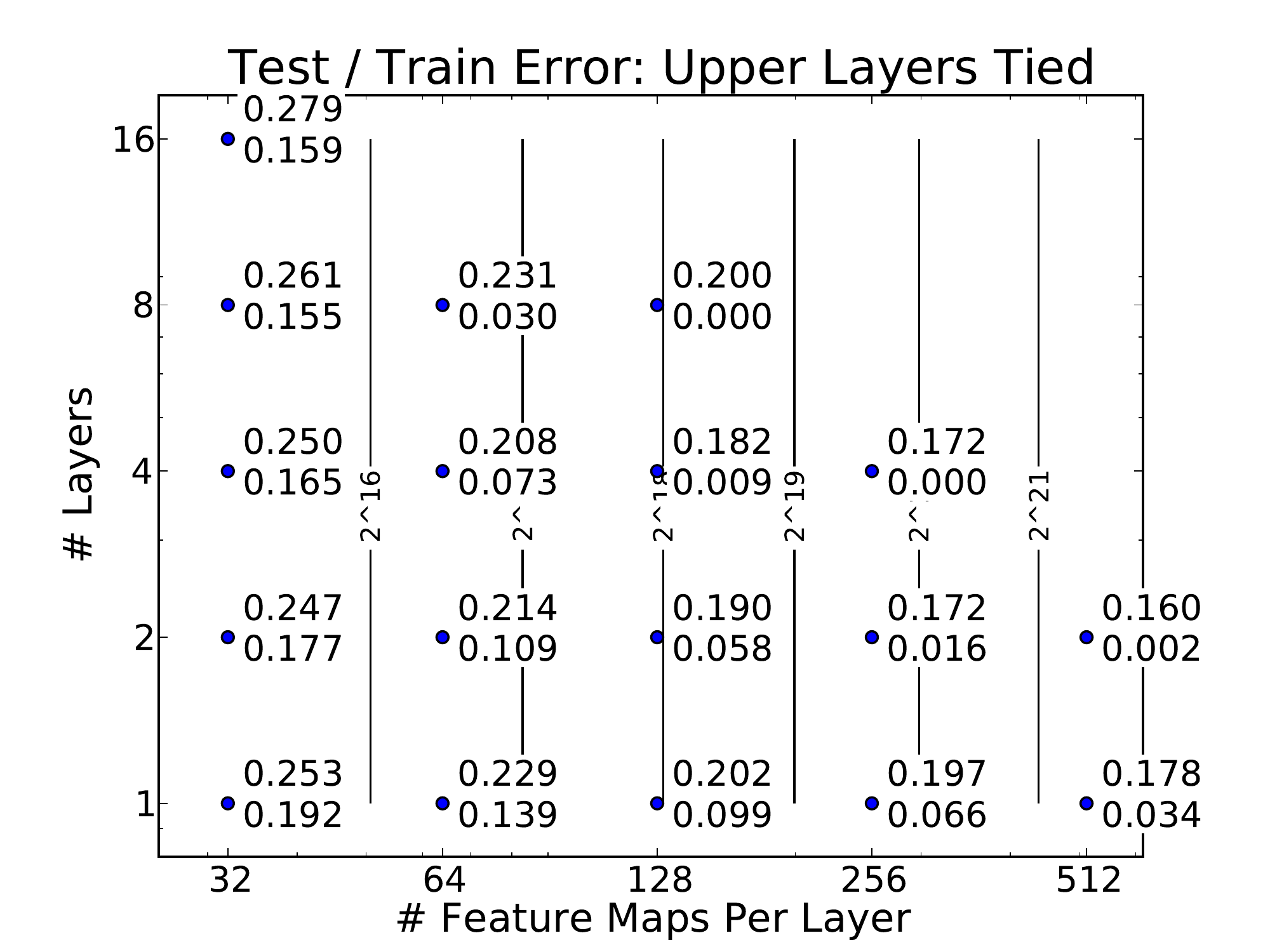}
\hspace{-1.45in}
\caption{Classification performance on CIFAR-10 as a function of network size, for untied (left) and tied (right) models.  Contours indicate lines along which the total number of parameters remains constant.  The top number by each point is test error, the bottom training error.}
\label{fig:cifar_nparams_nlayers}
\end{figure}

\begin{figure}[h]
\centering
\hspace{-1.5in}
\includegraphics[width=\plotwidth]{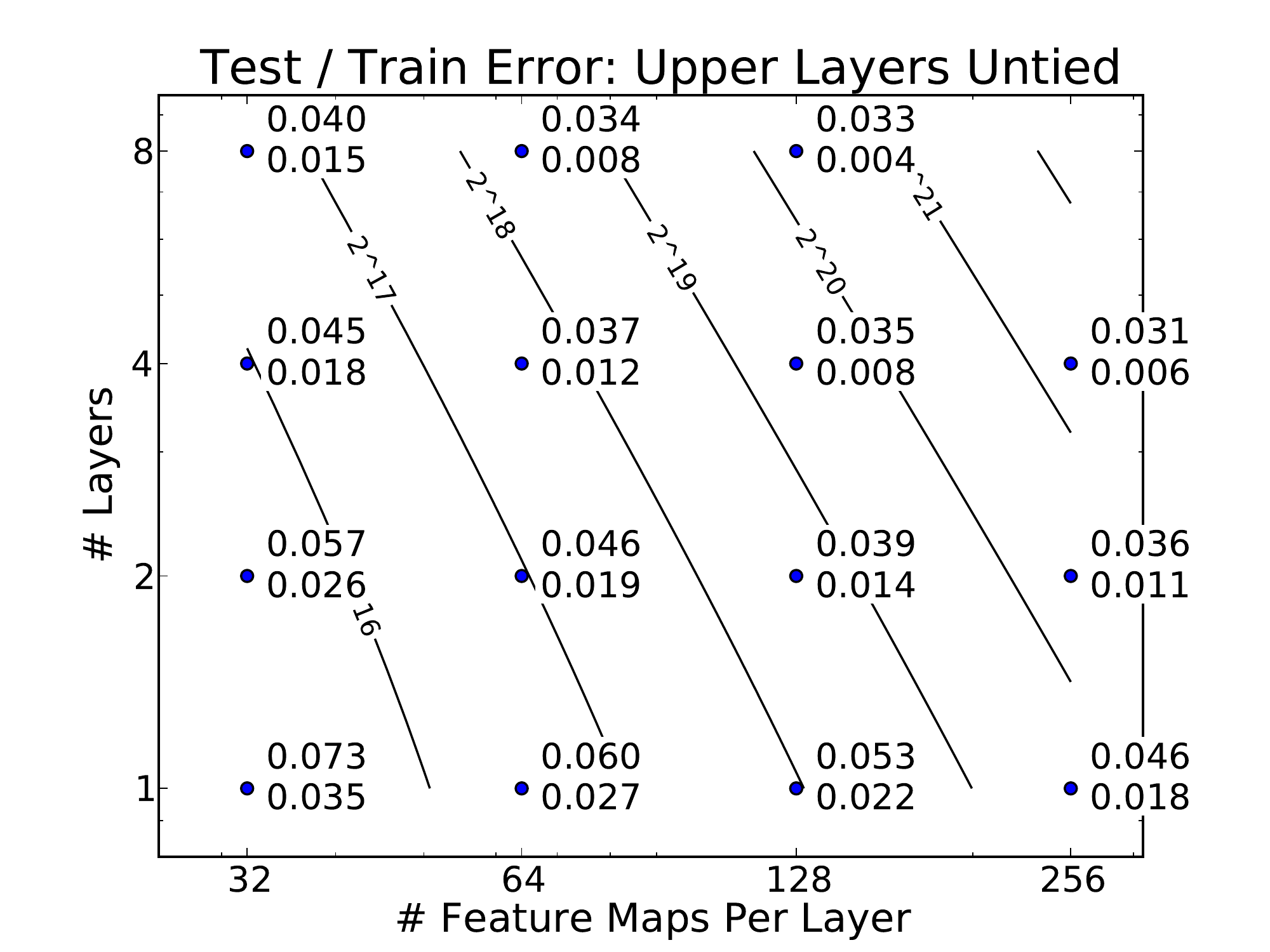}
\hspace{-0.1 in}
\includegraphics[width=\plotwidth]{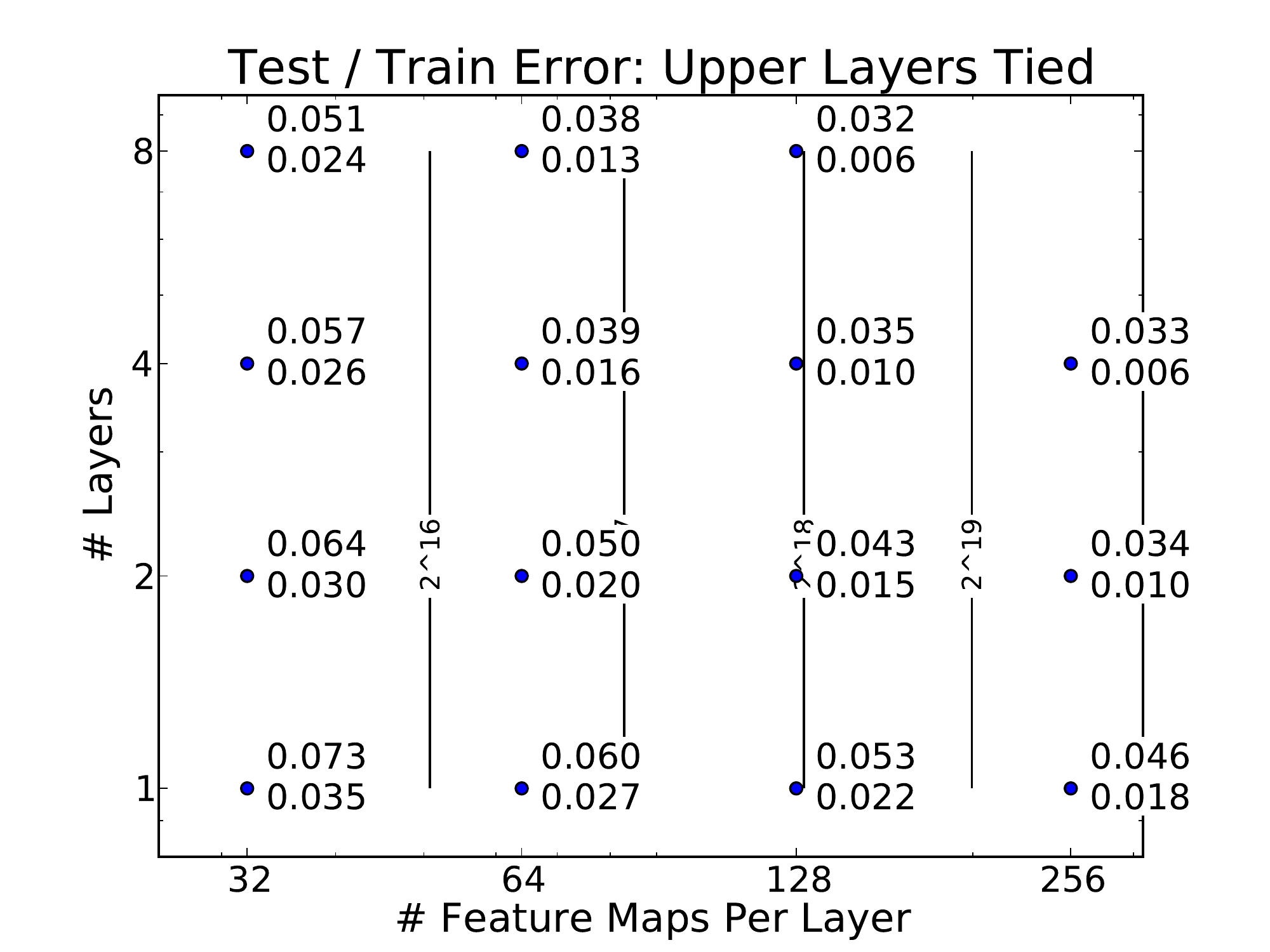}
\hspace{-1.45in}
\caption{Classification performance on Street View House Numbers as a function of network size, for untied (left) and tied (right) models.
}
\label{fig:svhn_nparams_nlayers}
\end{figure}

Results are shown in Figs.~\ref{fig:cifar_nparams_nlayers} and
\ref{fig:svhn_nparams_nlayers}.  Here, we plot each condition on a grid
according to numbers of feature maps and layers.  To the right of each point, we
show the test error (top) and training error (bottom).  Contours show curves
with a constant number of parameters:  in the untied case, the number of
parameters is determined by the number of feature maps and layers, while in the
tied case it is determined solely by the number of feature maps; \secc{case1} examines the behavior along these lines in more detail.

First, we note that despite the simple architecture of our model,
it still achieves competitive performance on both
datasets, relative to other models that, like ours, do not use any image transformations or other
regularizations such as dropout \cite{Hinton12,dropconnect},
stochastic pooling \cite{stochpool} or maxout \cite{maxout} (see \tab{compare}).
Thus our simplifications do not entail a departure from current
methods in terms of performance.

We also see roughly how the numbers of layers, feature maps and parameters
affect performance of these models at this range. In particular, increasing any of them
tends to improve performance, particularly on the training set (a notable
exception to CIFAR-10 at 16 layers in the tied case, which goes up
slightly).  We now examine the independent effects of each of these variables
in detail.

\vspace{-1mm}

\begin{table}[h]
\centering
\small
\begin{tabular}{|l|c|}
\hline 
\bf CIFAR-10  & \bf Test error (\%)\\
\hline \hline
Ours   & 16.0     \\ \hline
Snoek \etal \cite{Snoek}  & 15.0 \\
Ciresan \etal \cite{Ciresan11} & 15.9 \\
Hinton \etal \cite{Hinton12} & 16.6   \\
Coates \& Ng \cite{Coates2011} & 18.5 \\
\hline 
\end{tabular}
\quad \quad
\begin{tabular}{|l|c|}
\hline 
\bf SVHN  & \bf Test error (\%)\\
\hline \hline 
Ours   & 3.1     \\ \hline
Zeiler \& Fergus (max pool) \cite{stochpool} & 3.8 \\
Sermanet \etal \cite{Sermanet11} & 4.9 \\
\hline 
\end{tabular}
\caption{Comparison of our largest model architecture (measured by number of parameters)
  against other approaches
  that do not use data transformations or stochastic regularization
  methods.}
\label{tab:compare}
\vspace{-2mm}
\end{table}

\vspace{-2mm}
\subsection{Effects of the Numbers of Feature maps, Parameters and Layers}
\vspace{-1mm}

In a traditional
untied convolutional network, the number of feature maps $M$, layers $L$ and
parameters $P$ are interrelated:  Increasing the number of feature maps or layers
increases the total number of parameters in addition to the representational power
gained by higher dimensionality (more feature maps) or greater nonlinearity
(more layers).  But by using the tied version of our model, we can investigate
the effects of each of these three variables independently.

To accomplish this, we consider the following three cases, each of which we
investigate with the described setup:

\begin{enumerate}

\item  {\it Control for $M$ and $P$, vary $L$}:  Using the tied model (constant
$M$ and $P$), we evaluate performance for different numbers of layers $L$.

\item  {\it Control for $M$ and $L$, vary $P$}:  Compare pairs of tied and
untied models with the same numbers of feature maps $M$ and layers $L$.  The
number of parameters $P$ increases when going from tied to untied model for
each pair.

\item  {\it Control for $P$ and $L$, vary $M$}:  Compare pairs of untied and
tied models with the same number of parameters $P$ and layers $L$.  The number
of feature maps $M$ increases when going from the untied to tied model for each pair.

\end{enumerate}

Note the number of parameters $P$ is equal to the total number
of independent weights and biases over all layers, including initial feature
extraction and classification.  This is given by the
formula below for the untied model (for the tied case, substitute $L=1$ regardless of the number of layers):
\begin{equation*}
\begin{array}{cccccccccccc}
P & = & 8 \cdot 8 \cdot 3 \cdot M 
  & + & 3\cdot3\cdot M^2 \cdot L   
  & + & M \cdot (L + 1)            
  & + & 64 \cdot M \cdot 10 + 10
\\
&& \mbox{\small (first layer)}
&& \mbox{\small (higher layers)}
&& \mbox{\small (biases)}
&& \mbox{\small (classifier)}
\end{array}
\end{equation*}

\setlength{\plotwidth}{2.2in}

\subsubsection{Case 1: Number of Layers}
\label{sec:case1}

\begin{figure}[p]
\centering
\begin{tabular}{ccc}
\hline
\multicolumn{3}{c}{\bf Experiment 1a:  Error by Layers and Features (tied model)} \\
\hline
\parbox{0.5in}{\vspace{-0.75\plotwidth} \centering Test \\ Error} &
\includegraphics[width=\plotwidth]{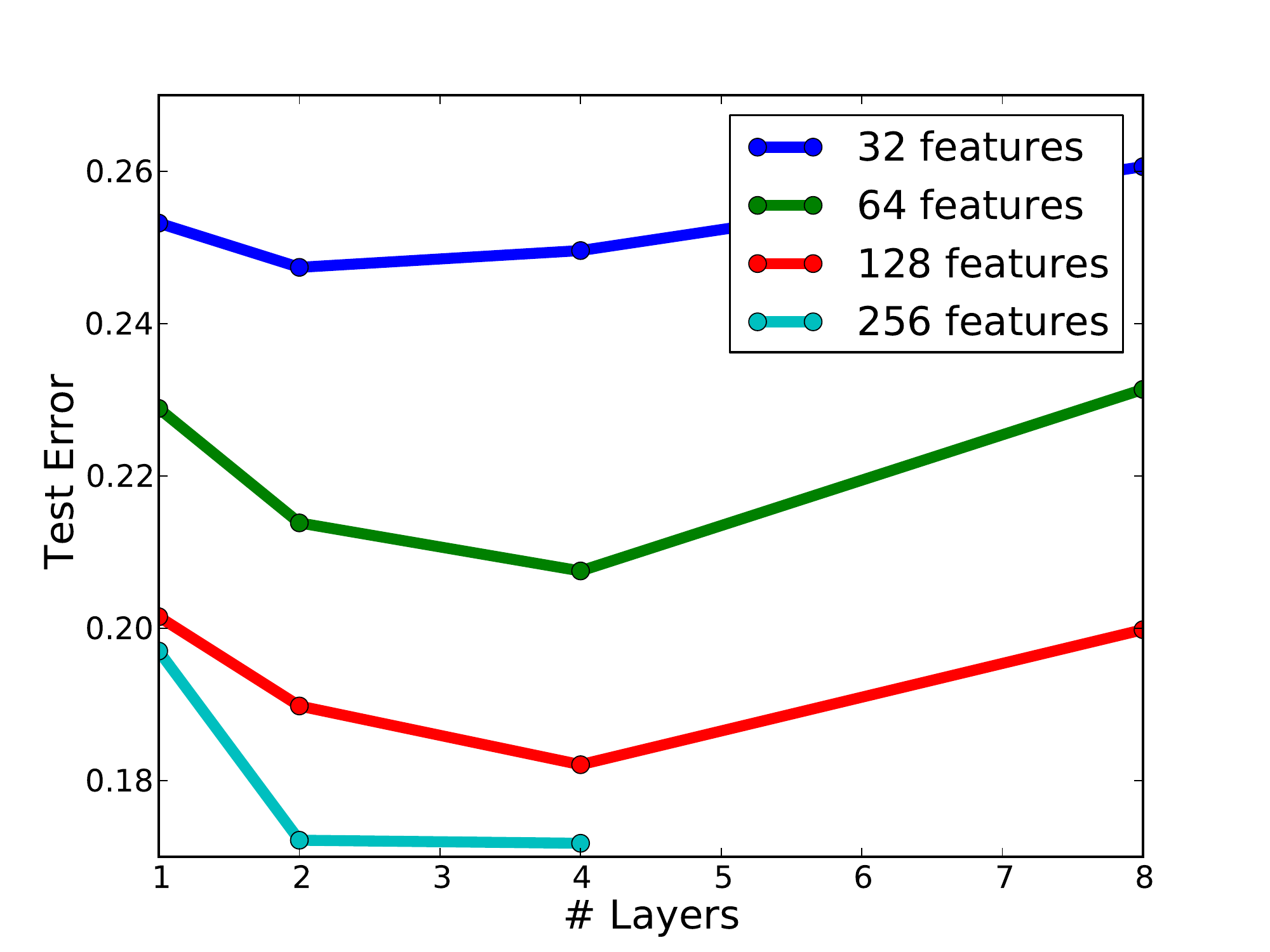}
&
\includegraphics[width=\plotwidth]{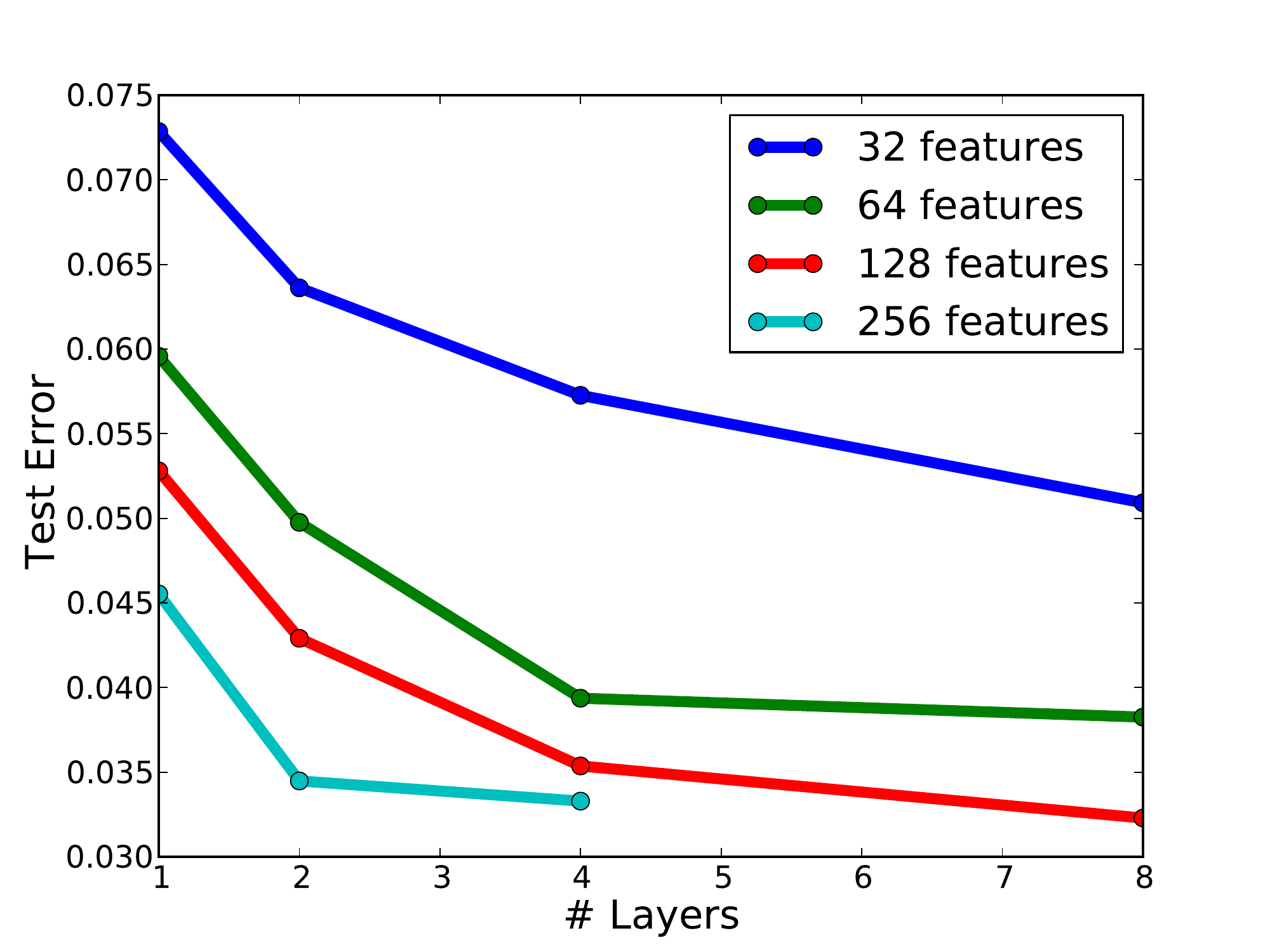}
\\
\parbox{0.5in}{\vspace{-0.75\plotwidth} \centering Training \\ Error} &
\includegraphics[width=\plotwidth]{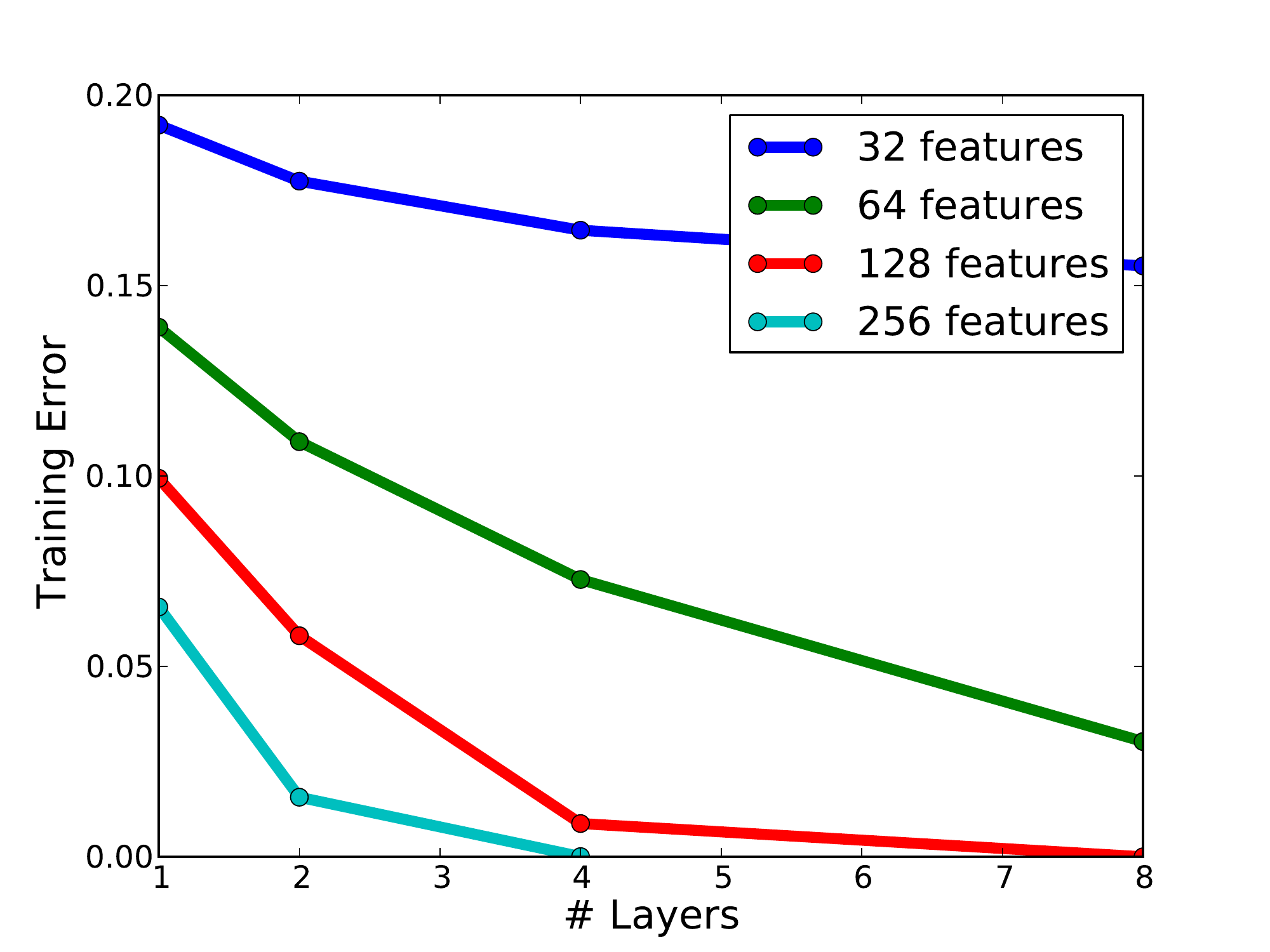}
&
\includegraphics[width=\plotwidth]{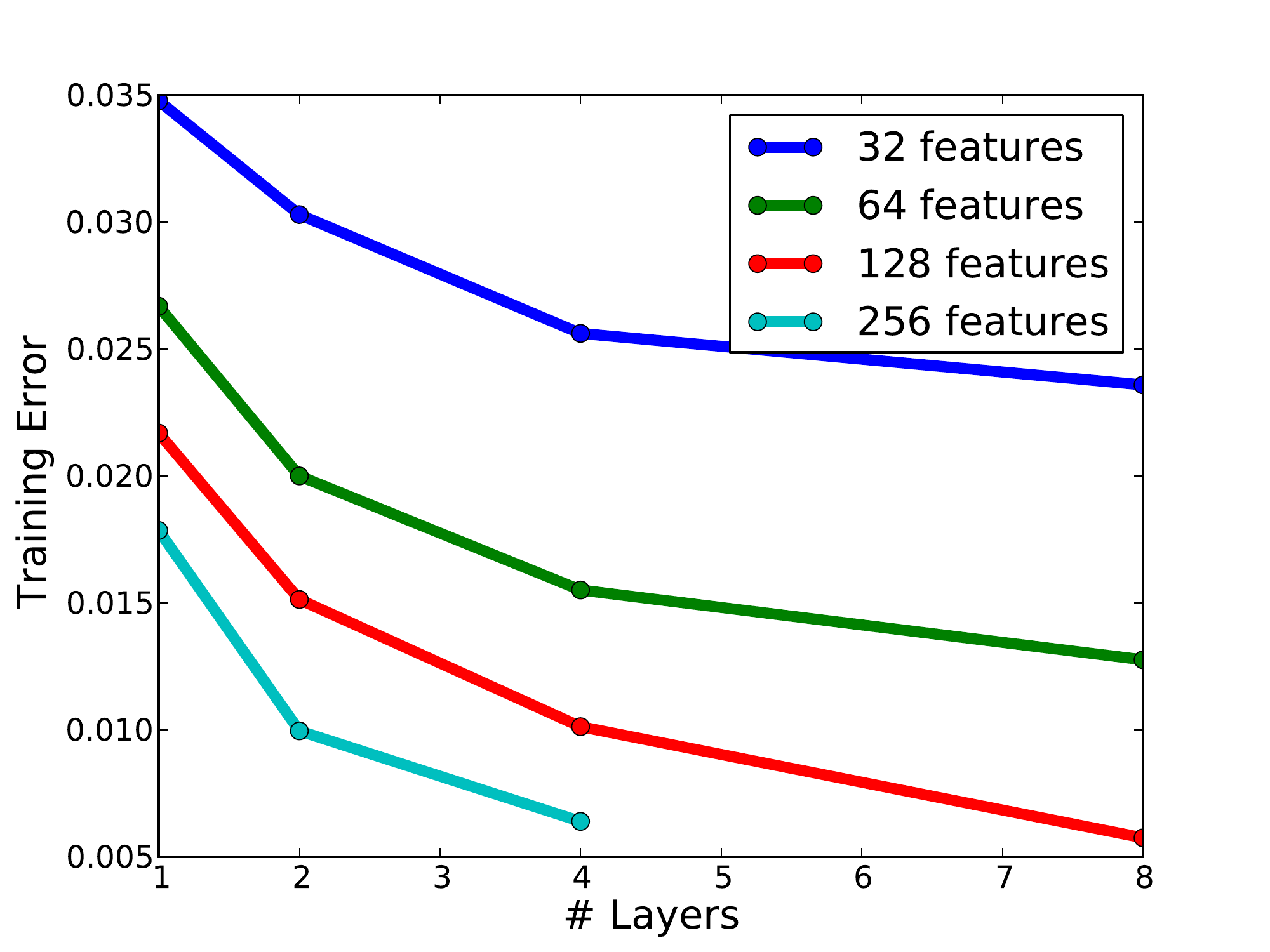}
\\
& (a) CIFAR-10 & (b) SVHN
\end{tabular}

\caption{ Comparison of classification error for different numbers of layers in
the tied model.  This controls for the number of parameters and features.  We
show results for both (a) CIFAR-10 and (b) SVHN datasets.  } 

\label{fig:fix_mp}
\end{figure}

We examine the first of these cases in \fig{fix_mp}.  Here, we plot
classification performance at different numbers of layers using the tied model
only, which controls for the number of parameters.  A different curve is shown
for different numbers of feature maps.  For both CIFAR-10 and SVHN, performance
gets better as the number of layers increases, although there is an upward tick
at 8 layers for CIFAR-10 test error.
The predominant cause of
this appears to be overfitting, since the training error still goes down.
At these depths, therefore, adding more layers alone tends to increase performance, even though no additional parameters are introduced.
This is because additional layers allow the network to learn more complex
functions by using more nonlinearities. 

\begin{figure}[p]
\centering
\begin{tabular}{ccc}
\hline
\multicolumn{3}{c}{\bf Experiment 1b:  Error by Parameters and Layers (untied model)} \\
\hline
\parbox{0.5in}{\vspace{-0.75\plotwidth} \centering Test \\ Error} &
\includegraphics[width=\plotwidth]{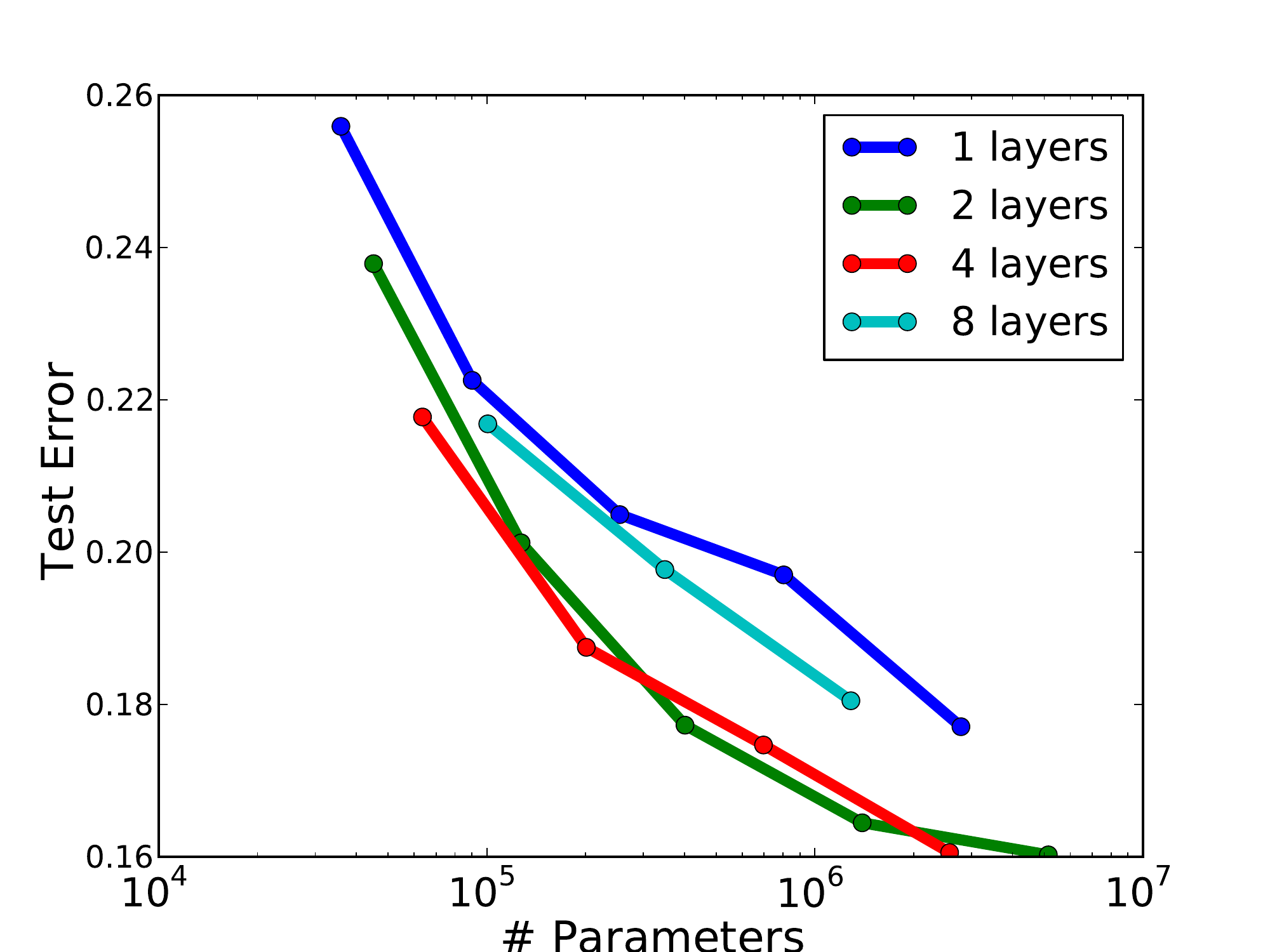}
&
\includegraphics[width=\plotwidth]{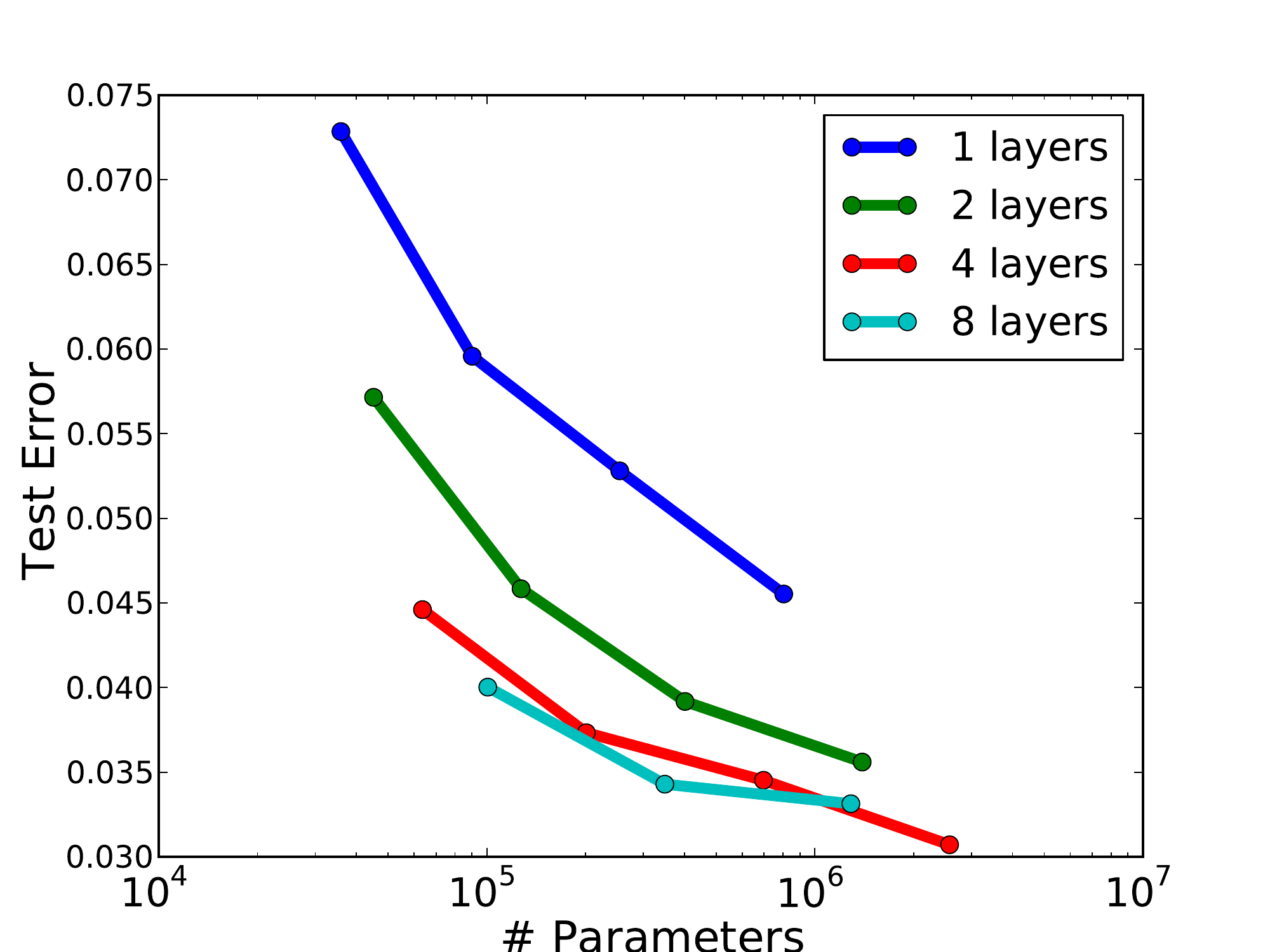}
\\
\parbox{0.5in}{\vspace{-0.75\plotwidth} \centering Training \\ Error} &
\includegraphics[width=\plotwidth]{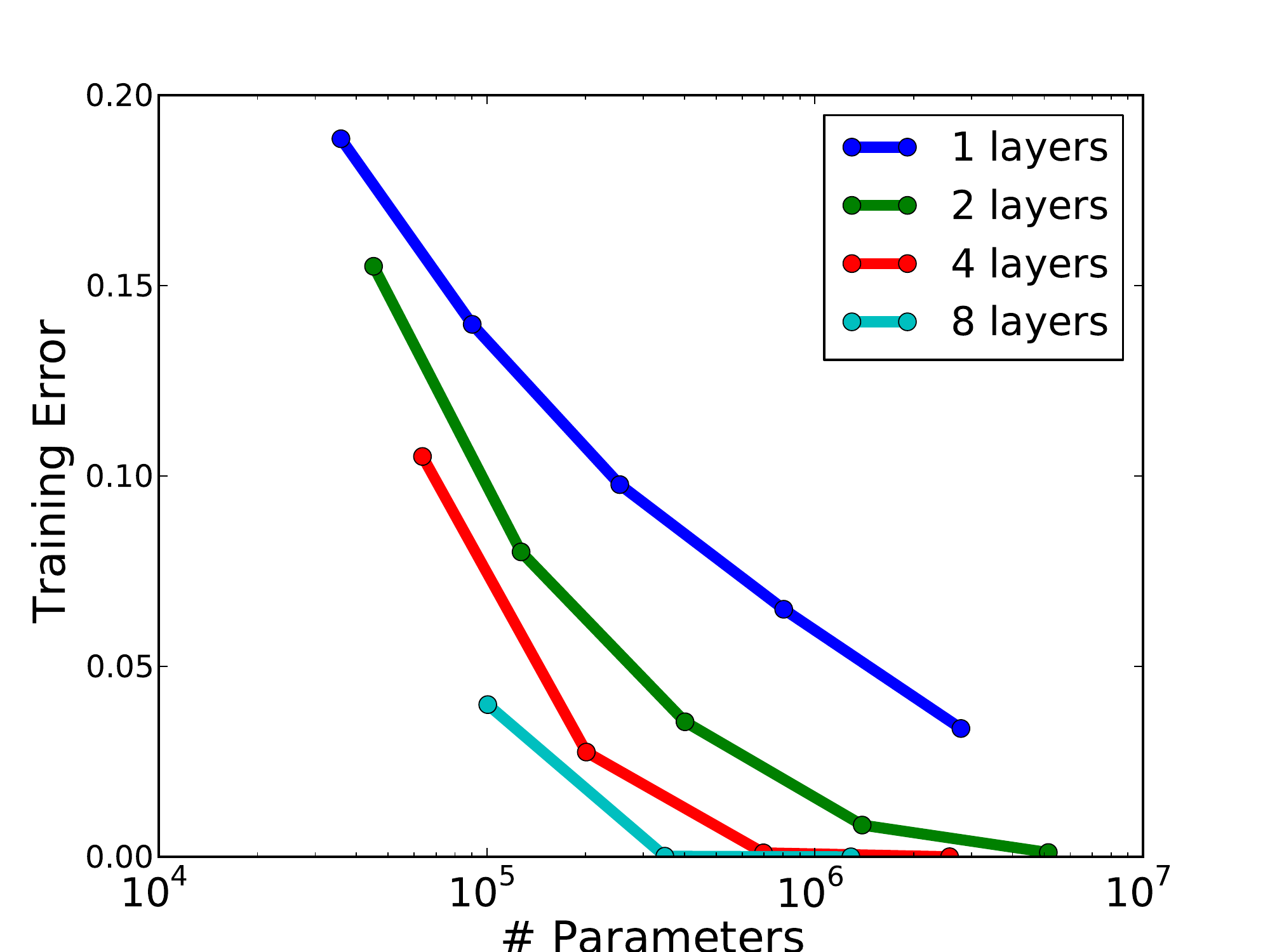}
&
\includegraphics[width=\plotwidth]{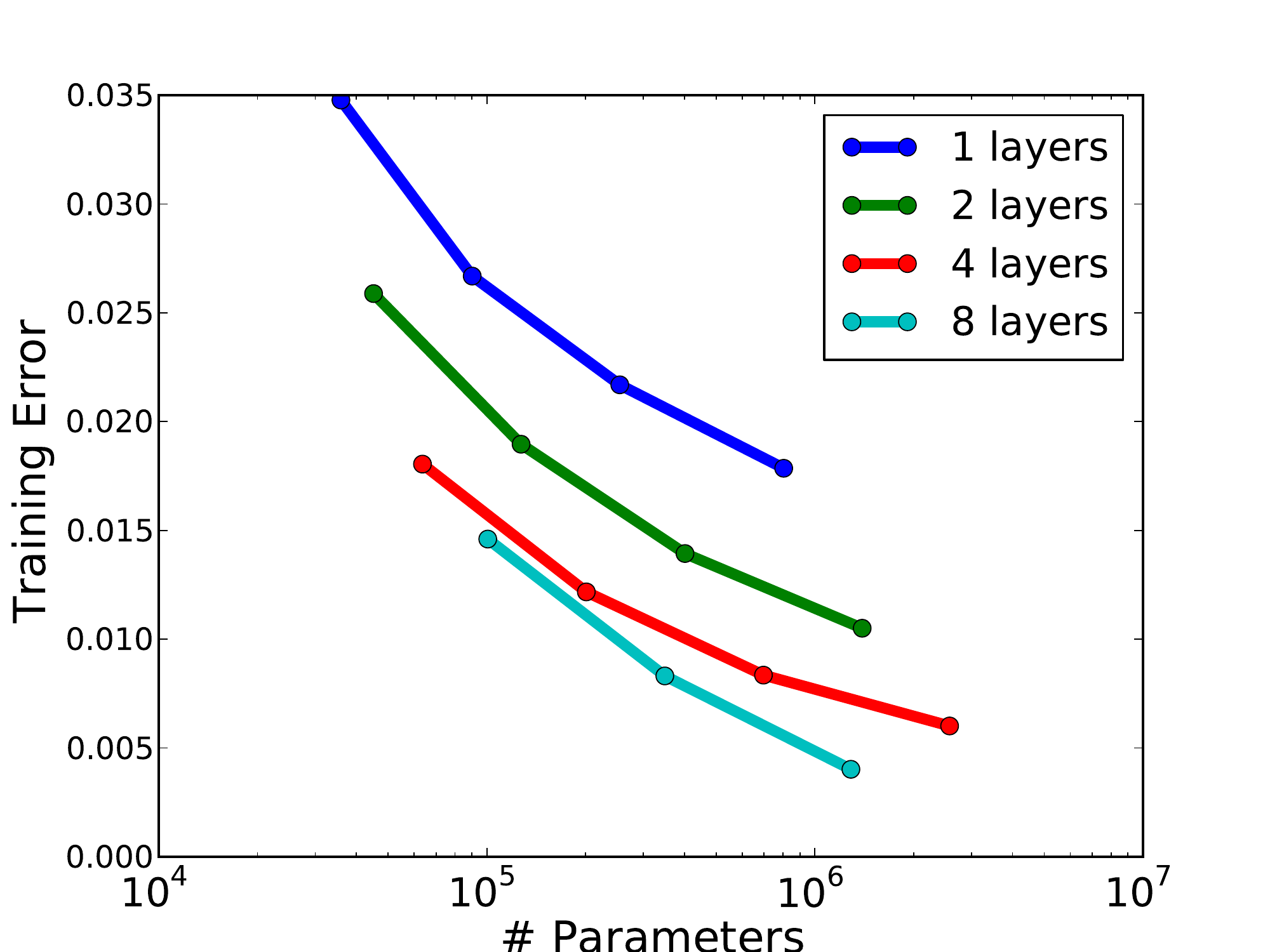}
\\
& (a) CIFAR-10 & (b) SVHN
\end{tabular}

\caption{ Comparison of classification error for different numbers of parameters in
the untied model, for (a) CIFAR-10 and (b) SVHN datasets.  
Larger numbers of both parameters and layers help performance.  In addition,
for a fixed budget of parameters, allocating them in more layers is generally better
(CIFAR-10 test error increases above 4 layers due to overfitting).
} 
\vspace{-2mm}
\label{fig:untied_pl}
\end{figure}

This conclusion is further supported by \fig{untied_pl}, which shows
performance of the untied model according to numbers of parameters and layers.
Note that vertical cross-sections of this figure correspond to the
constant-parameter contours of \fig{cifar_nparams_nlayers}.  Here, we can also
see that for any given number of parameters, the best performance is obtained
with a deeper model. The exception to this is again the 8-layer models on
CIFAR-10 test error, which suffer from overfitting.

\subsubsection{Case 2: Number of Parameters}

To vary the number of parameters $P$ while holding fixed the number of feature
maps $M$ and layers~$L$, we consider pairs of tied and untied models where $M$
and $L$ remain the same within each pair.  The number of parameters $P$ is then
greater for the untied model.

The result of this comparison is shown in \fig{fix_ml}.  Each point corresponds
to a model pair; we show classification performance of the tied model on the
$x$ axis, and performance of the untied model on the $y$ axis.  Since the
points fall below the $y=x$ line, classification performance is better for the
untied model than it is for the tied.  This is not surprising, since the untied
model has more total parameters and thus more flexibility.  Note also that the
two models converge to the same test performance as classification gets better
--- this is because for the largest numbers of $L$ and $M$, both models have
enough flexibility to achieve maximum test performance and begin to overfit.

\setlength{\plotwidth}{2.0in}

\begin{figure}[h]
\centering
\begin{tabular}{ccc}
\hline
\multicolumn{3}{c}{\bf Experiment 2:  Same Feature Maps and Layers, Varied Parameters} \\
\hline
&&\\
\parbox{0.5in}{\vspace{-0.9\plotwidth} \centering Test \\ Error} &
\includegraphics[width=\plotwidth]{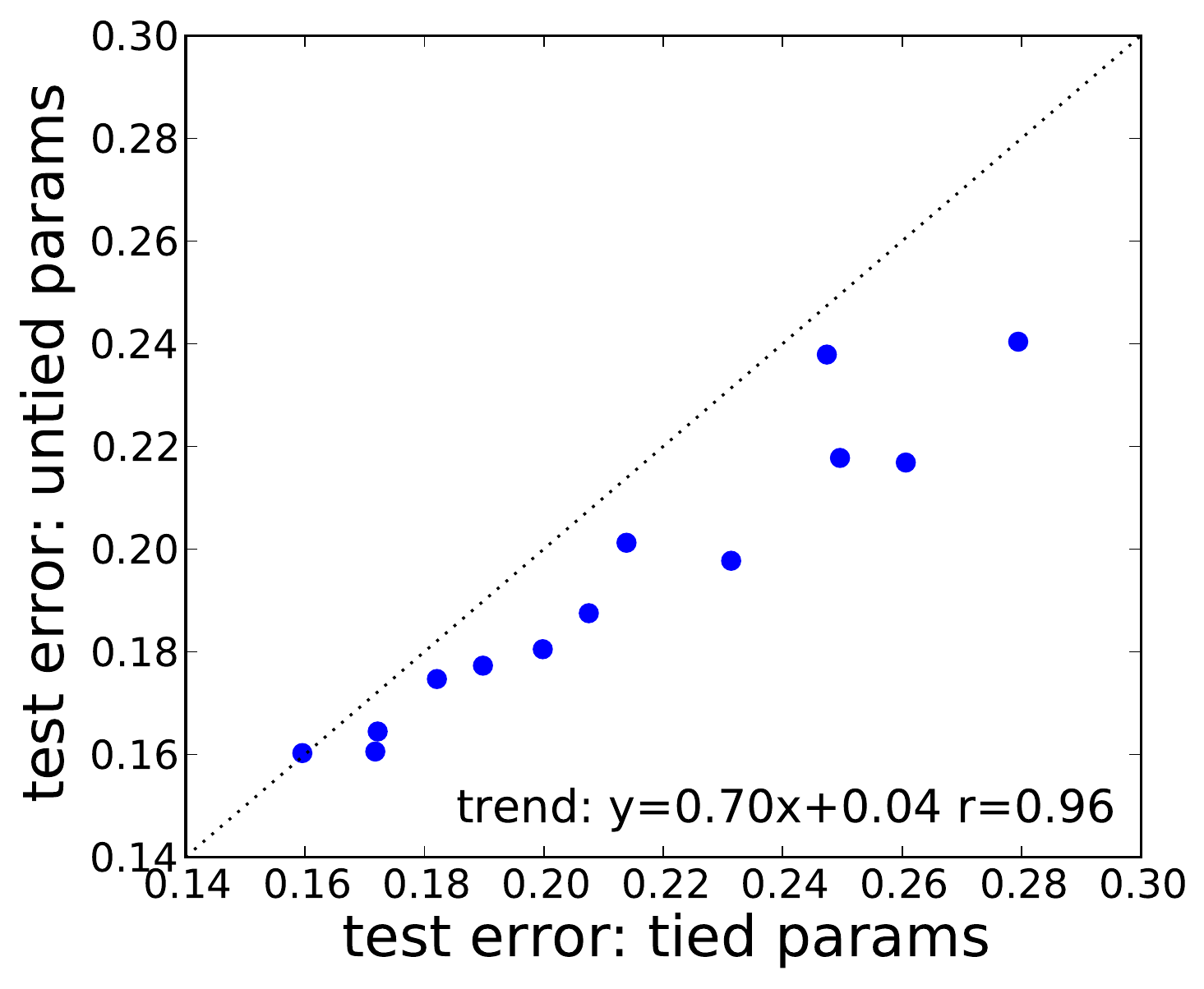}
&
\includegraphics[width=\plotwidth]{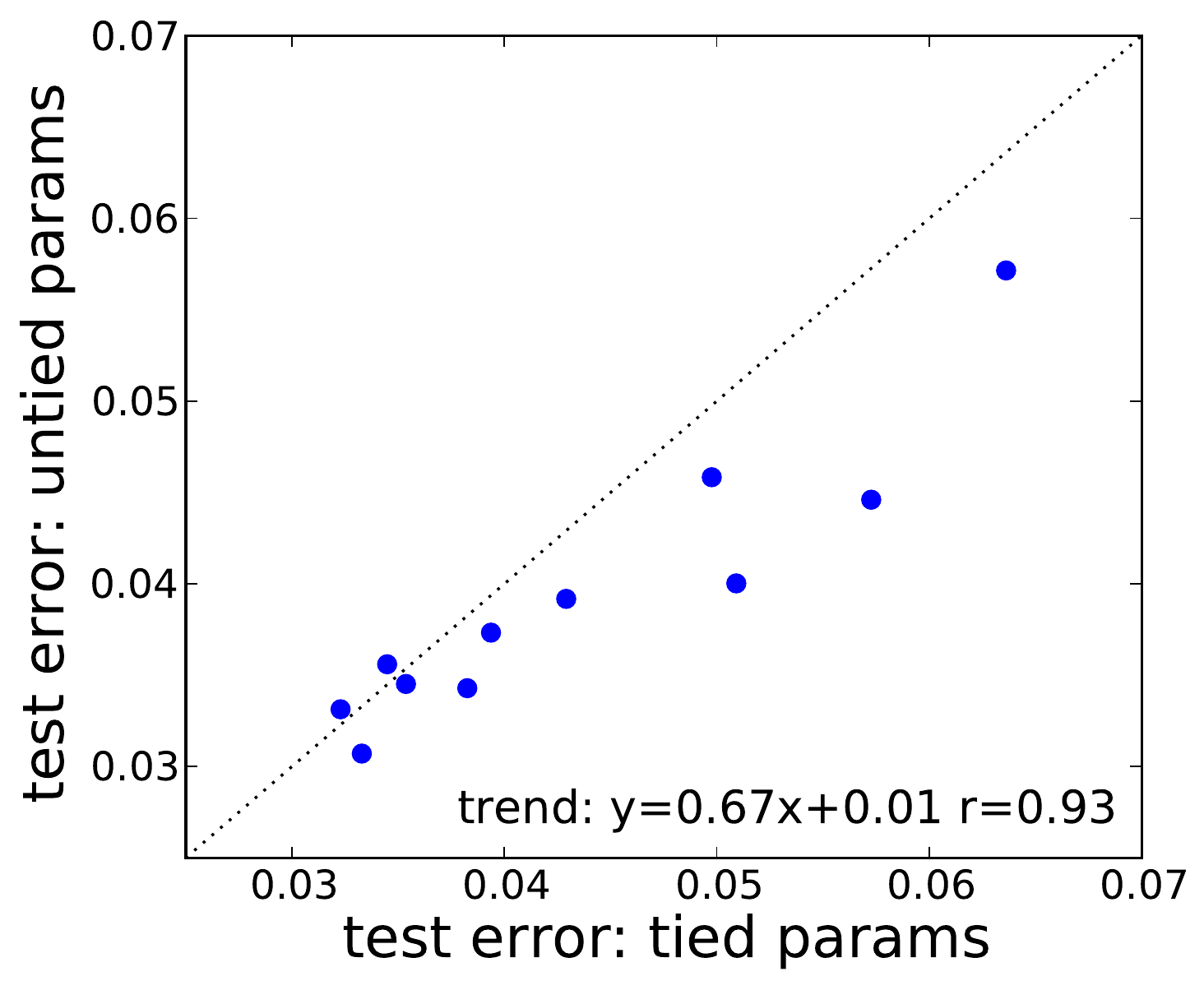}
\\
\parbox{0.5in}{\vspace{-0.9\plotwidth} \centering Training \\ Error} &
\includegraphics[width=\plotwidth]{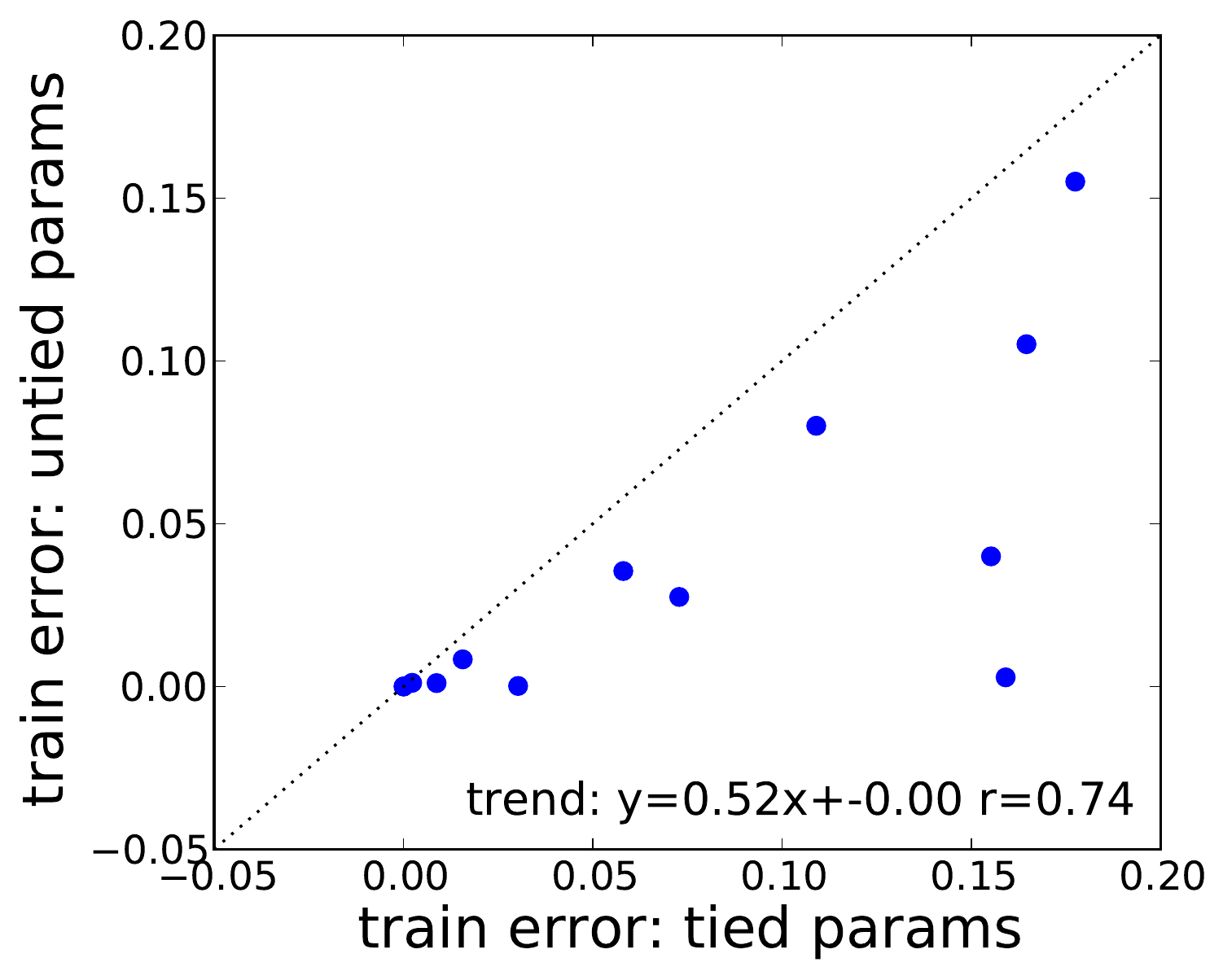}
&
\includegraphics[width=\plotwidth]{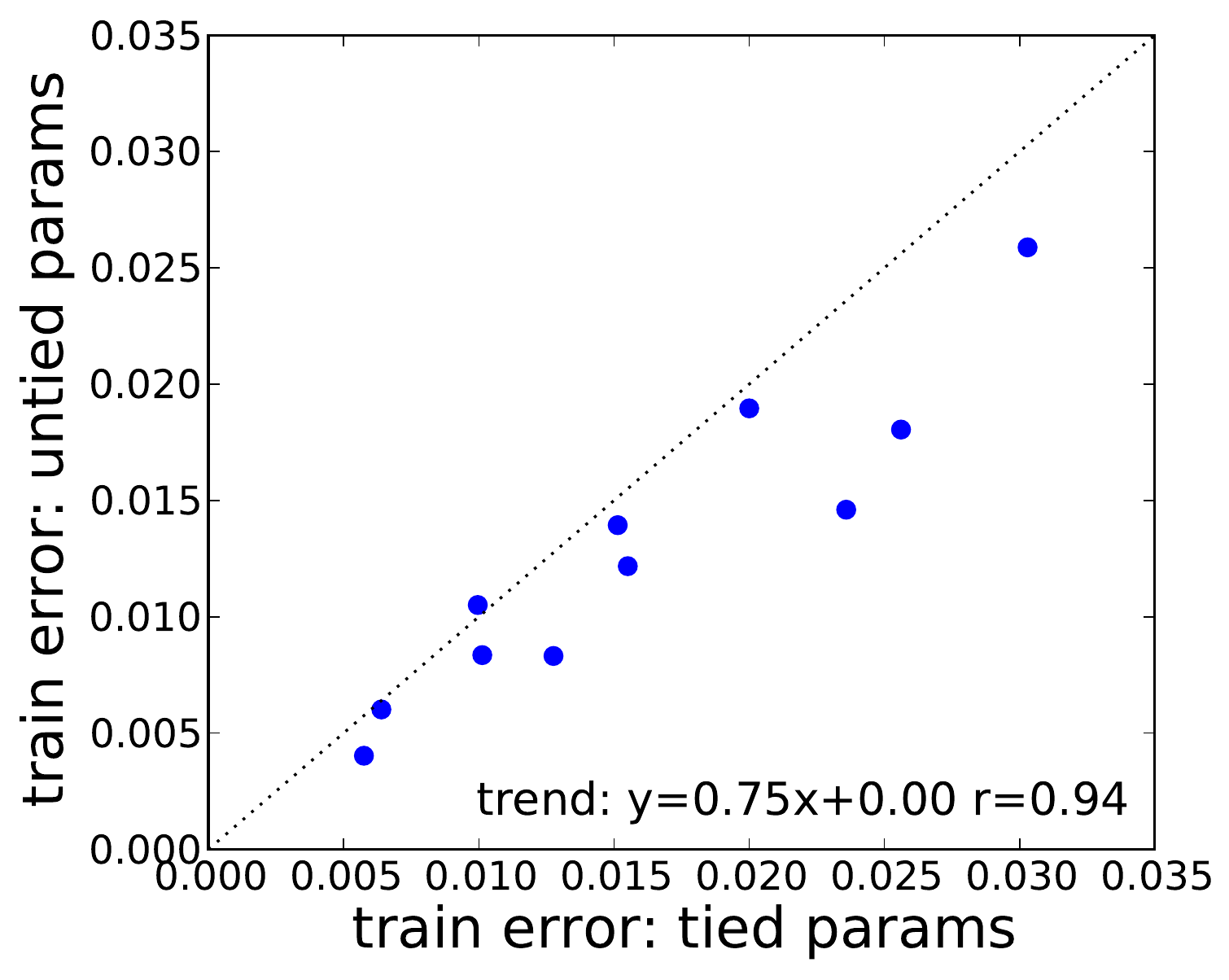}
\\
& (a) CIFAR-10 & (b) SVHN \\
\end{tabular}

\caption{ Comparison of classification error between tied and untied models,
controlling for the number of feature maps and layers.
Linear regression coefficients are in the bottom-right
corners.
} 

\label{fig:fix_ml}
\vspace{-3mm}
\end{figure}

\subsubsection{Case 3: Number of Feature Maps}

We now consider the third condition from above, the effect of varying the
number of feature maps~$M$ while holding fixed the numbers of layers $L$ and
parameters $P$.

For a given $L$,
we find model pairs whose numbers of parameters $P$ are very close by varying
the number of feature maps.  For example, an untied model with $L=3$ layers and $M=71$ feature maps has $P=195473$ parameters,
while a tied model with $L=3$ layers and $M=108$ feature maps has $P=195058$ parameters --- a difference of
only 0.2\%.
In this experiment, we randomly sampled model pairs having
the same number of layers, and where the numbers of parameters were within 1.0\%
of each other.  We considered models where the number of layers beyond the
first was between 2 and 8, and the number of feature maps was between 16 and
256 (for CIFAR-10) or between 16 and 150 (for SVHN).

\begin{figure}[t]
\centering
\begin{tabular}{ccc}
\hline
\multicolumn{3}{c}{\bf Experiment 3:  Same Parameters and Layers, Varied Feature Maps} \\
\hline
\parbox{0.5in}{\vspace{-0.9\plotwidth} \centering Test \\ Error} &
\includegraphics[width=\plotwidth]{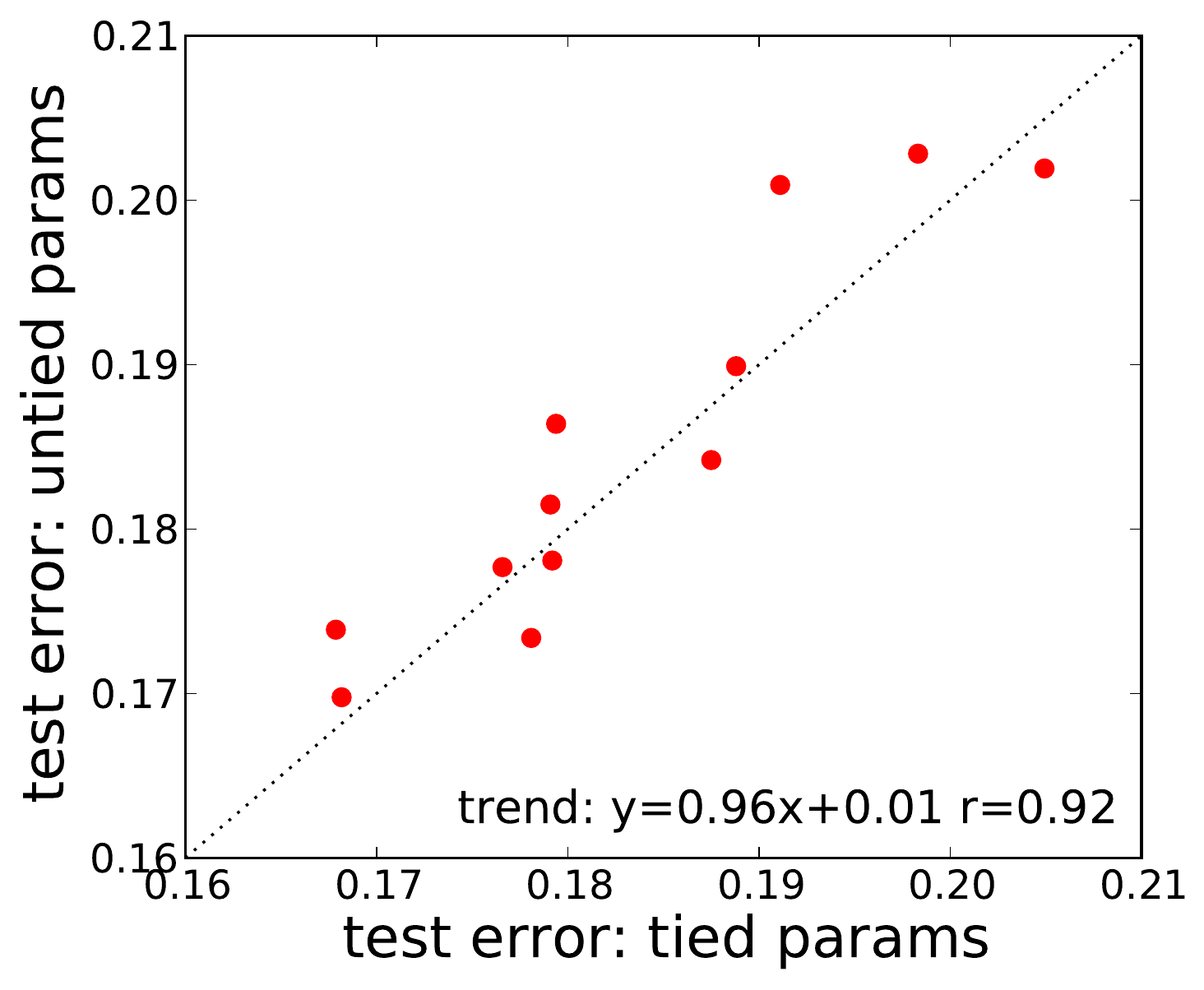}
&
\includegraphics[width=\plotwidth]{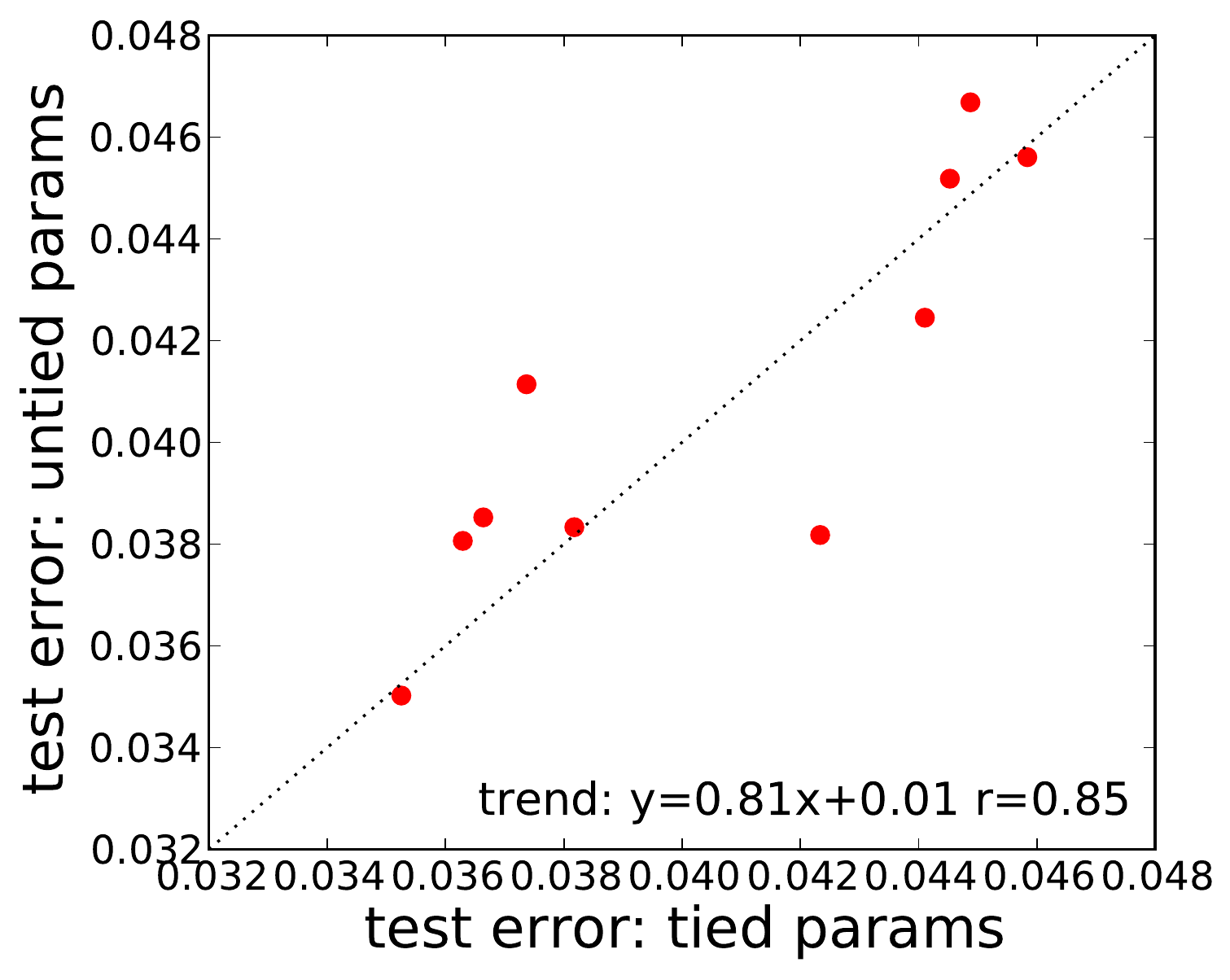}
\\
\parbox{0.5in}{\vspace{-0.9\plotwidth} \centering Training \\ Error} &
\includegraphics[width=\plotwidth]{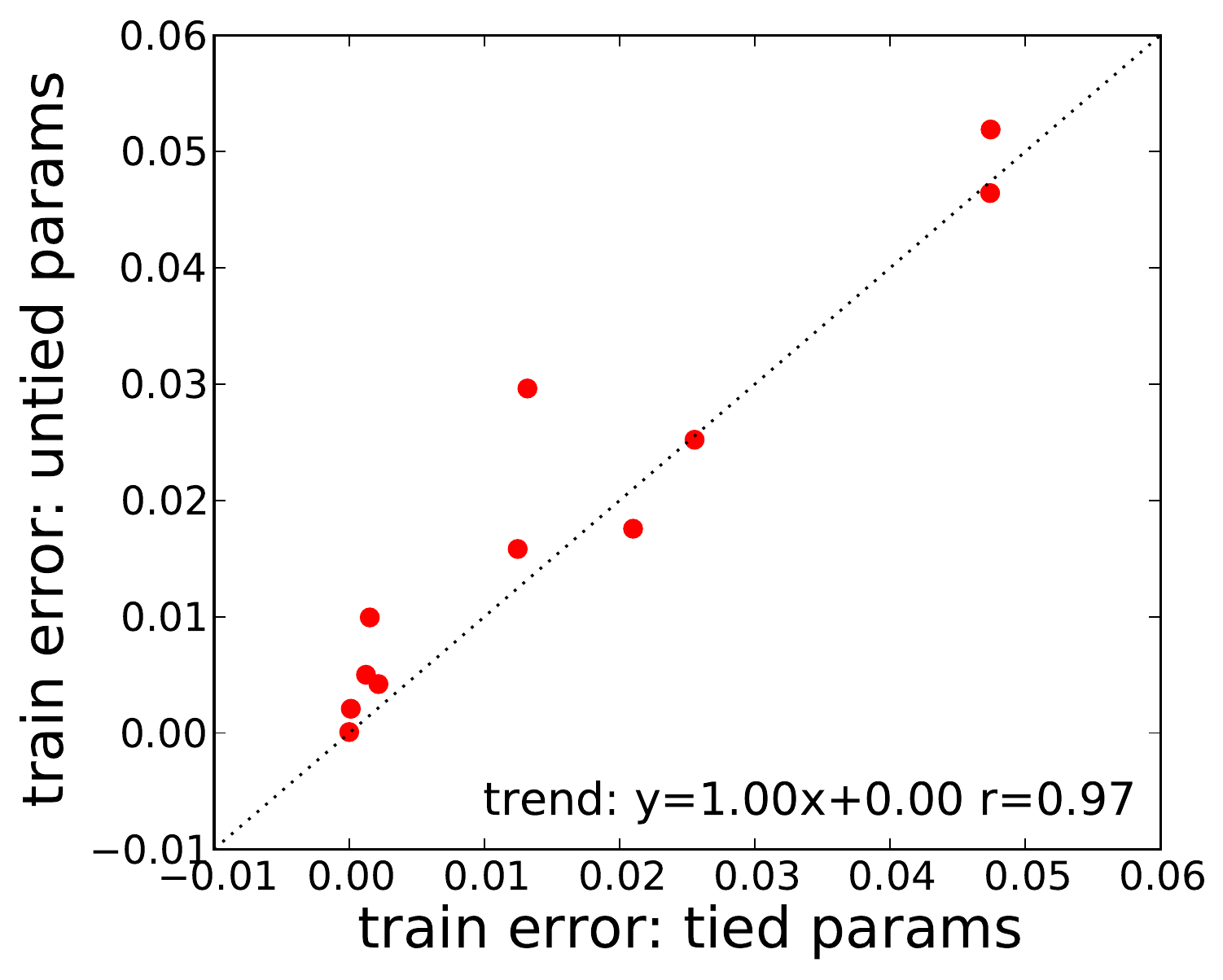}
&
\includegraphics[width=\plotwidth]{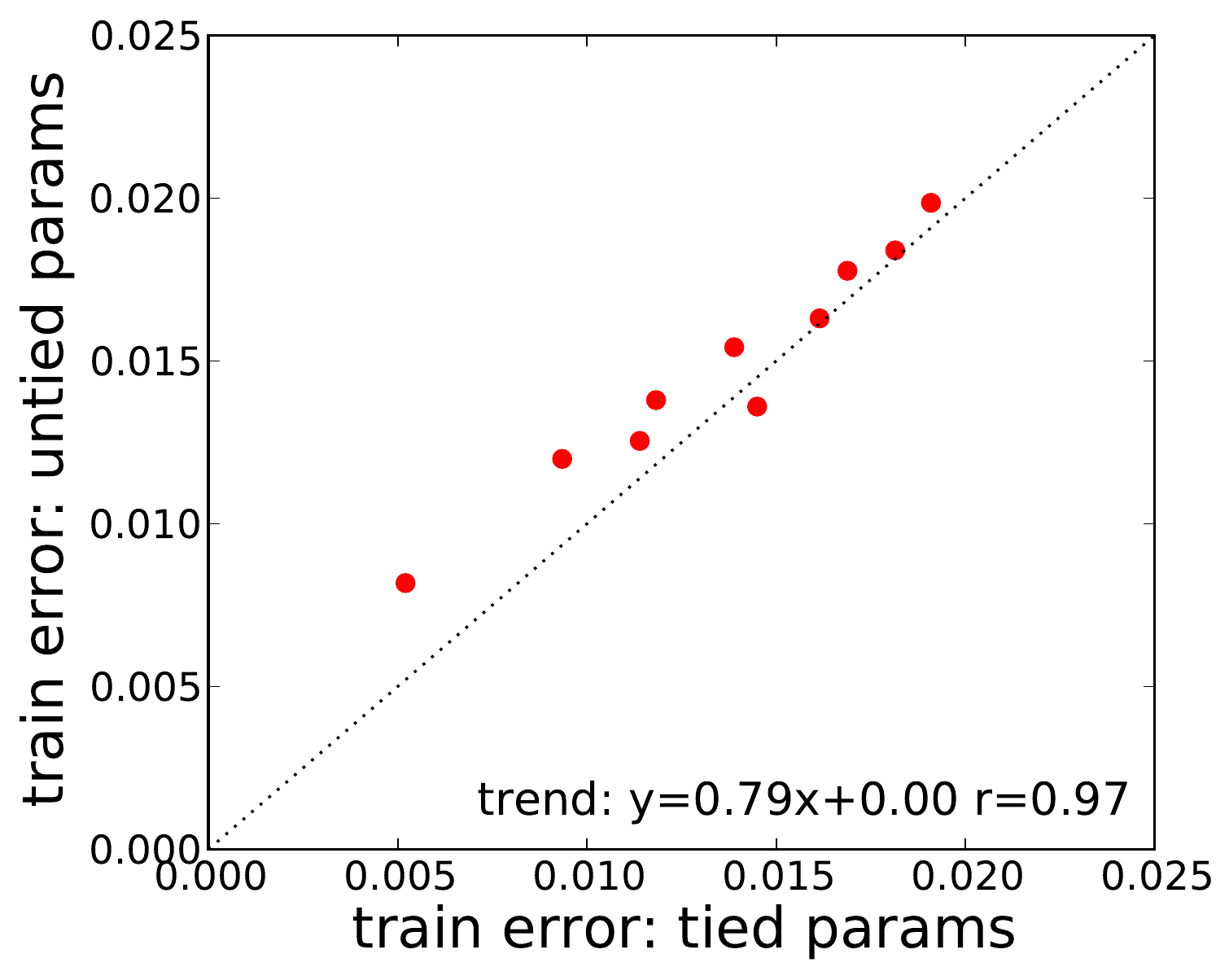}
\\
& (a) CIFAR-10 & (b) SVHN
\end{tabular}

\caption{ Comparison of classification error between tied and untied models,
controlling for the number of parameters and layers.
 Linear regression coefficients in the bottom-right corners.
} 

\vspace{-3mm}
\label{fig:fix_pl}
\end{figure}

\fig{fix_pl} shows the results.  As before, we plot a point for each model
pair, showing classification performance of the tied model on the $x$ axis, and
of the untied model on the $y$ axis.  This time, however, each pair has fixed
$P$ and $L$, and tied and untied models differ in their number of feature maps
$M$.  We find that despite the different numbers of feature maps, the tied and
untied models perform about the same in each case.  Thus, performance is
determined by the number of parameters and layers, and is insensitive to the
number of feature maps.

\section{Discussion}

Above we have demonstrated that while the numbers of layers and parameters each
have clear effects on performance, the number of feature maps has little
effect, once the number of parameters is taken into account.  This is perhaps
somewhat counterintuitive, as we might have expected the use of
higher-dimensional representations to increase performance; instead we find
that convolutional layers are insensitive to this size.

This observation is also consistent with \fig{untied_pl}:  Allocating a fixed
number of parameters across multiple layers tends to increase performance
compared to putting them in few layers, even though this comes at the cost of
decreasing the feature map dimension.  This is precicesly what one might expect
if the number of feature maps had little effect compared to the number of
layers.

Our analysis employed a special tied architecture and comes with some important
caveats, however. 
First, while the tied architecture serves as a useful point of comparison
leading to several interesting conclusions, it is new and thus its behaviors
are still relatively unknown compared to the common untied
models.  This may particularly apply to models with a large number of layers
($L > 8$), or very small numbers of feature maps ($M < 16$), which have been
left mostly unexamined in this paper.
Second, our experiments all used a simplified architecture, with
just one layer of pooling.  While we believe the principles found in
our experiments are likely to apply in more complex cases as well, this is
unclear and requires further investigation to confirm.

The results we have presented provide empirical confirmation within the context
of convolutional layers that increasing layers alone can yield performance
benefits (Experiment 1a).  They also indicate that filter parameters may be
best allocated in multilayer stacks (Experiments 1b and 3), even at the expense
of having fewer feature maps.  In conjunction with this, we find the feature map
dimension itself has little effect on convolutional layers'
performance, with most sizing effects coming from the numbers of layers and
parameters (Experiments 2 and 3).  Thus, focus would be best placed on
these variables when determining model architectures.

{
\small
\clearpage
\baselineskip=2pt
\bibliographystyle{plain}
\bibliography{reconv}
}

\end{document}